\pdfoutput=1

\documentclass[11pt]{article}
     
\usepackage[]{acl}

\usepackage{times}
\usepackage{latexsym}

\usepackage[T1]{fontenc}

\usepackage[utf8]{inputenc}

\usepackage{microtype}

\usepackage{amsmath}
\usepackage{amsfonts}
\usepackage{booktabs}
\usepackage{graphicx}
\usepackage[normalem]{ulem}
\useunder{\uline}{\ul}{}
\usepackage{dsfont}
\usepackage{xspace}
\usepackage{hyperref}
\usepackage{subcaption}
\usepackage{cleveref}
\usepackage{bbm}

\DeclareMathOperator*{\argmin}{argmin}
\DeclareMathOperator*{\argmax}{argmax}

\newcommand{\proposeddist}{\mbox{RCMD}\xspace}
\newcommand{\sparsedist}{\mbox{RCMD\textsubscript{sparse}}\xspace}
\newcommand{\densedist}{\mbox{RCMD\textsubscript{dense}}\xspace}
\newcommand{\proposed}{\mbox{CL\proposeddist}\xspace}

\newcommand{\simcsecls}{SimCSE\textsubscript{cls}\xspace}
\newcommand{\simcseavg}{SimCSE\textsubscript{avg}\xspace}

\newcommand{\bertbase}{BERT\textsubscript{base}\xspace}
\newcommand{\robertabase}{RoBERTa\textsubscript{base}\xspace}

\title{Toward Interpretable Semantic Textual Similarity via Optimal~Transport-based Contrastive Sentence Learning}

\newcommand*\samethanks[1][\value{footnote}]{\footnotemark[#1]}

\author{Seonghyeon Lee\textsuperscript{1}\thanks{ \ \ This work was done during internship at Scatterlab.}\ , Dongha Lee\textsuperscript{2}, Seongbo Jang\textsuperscript{1}\samethanks\ , and Hwanjo Yu\textsuperscript{1}\thanks{ \ \ Corresponding author} \\
  Computer Science and Engineering, POSTECH, Pohang, Republic of Korea\textsuperscript{1} \\
  University of Illinois at Urbana-Champaign (UIUC), Urbana, IL, United States\textsuperscript{2} \\
  \texttt{\{sh0416,jang.sb,hwanjoyu\}@postech.ac.kr, donghal@illinois.edu} \\}

\begin{document}
\maketitle

\begin{abstract}
Recently, finetuning a pretrained language model to capture the similarity between sentence embeddings has shown the state-of-the-art performance on the semantic textual similarity (STS) task.
However, the absence of an interpretation method for the sentence similarity makes it difficult to explain the model output.
In this work, we explicitly describe the sentence distance as the weighted sum of contextualized token distances on the basis of a transportation problem, and then present the optimal transport-based distance measure, named \proposeddist;
it identifies and leverages semantically-aligned token pairs.
In the end, we propose \proposed, a contrastive learning framework that optimizes \proposeddist of sentence pairs, which enhances the quality of sentence similarity and their interpretation.
Extensive experiments demonstrate that our learning framework outperforms other baselines on both STS and interpretable-STS benchmarks, indicating that it computes effective sentence similarity and also provides interpretation consistent with human judgement.
The code and checkpoint are publicly available at \url{https://github.com/sh0416/clrcmd}.

\end{abstract}

\section{Introduction}
\label{sec:intro}
Predicting the semantic similarity between two sentences has been extensively studied in the literature \citep{gomaa2013survey,agirre-etal-2015-semeval,majumder2016semantic,cer-etal-2017-semeval}.
Several recent studies successfully utilized a pretrained language model such as BERT \citep{devlin-etal-2019-bert} by finetuning it to capture sentence similarity \citep{reimers-gurevych-2019-sentence}.
To be specific, they define a similarity score between sentence embeddings, which are obtained by aggregating contextualized token embeddings (e.g., avg pooling) or using a special token (e.g., \texttt{[CLS]}), then optimize the score for natural language inference (NLI) or semantic textual similarity (STS) tasks \citep{gao2021simcse}.

Along with the quality of sentence similarity, interpreting the predicted sentence similarity is also important for end-users to better understand the results \citep{agirre-etal-2016-semeval-2016,gilpin2018explaining,rogers-etal-2020-primer}.
In general, finding out the cross-sentence alignment and the importance of each aligned part is useful for analyzing sentence similarity \citep{sultan-etal-2015-dls}.
For example, there were several attempts to use explicit features (e.g., TF-IDF) for easily analyzing the interaction among the shared terms \citep{salton1988term} or to adopt sophisticated metrics (e.g., word mover's distance) for explicitly describing it by the importance and similarity of word pairs across two sentences \citep{pmlr-v37-kusnerb15}.
However, for recent approaches that leverage sentence embeddings from a pretrained model, it has not been studied how the cross-sentence interaction of each part contributes to the final sentence similarity.

\begin{figure}
\centering
\includegraphics[width=0.99\linewidth]{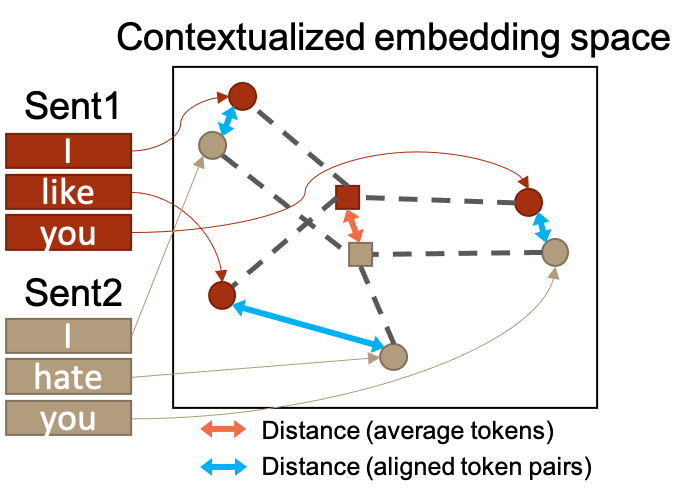}
\caption{An illustrative example of a transportation problem in contextualized embedding space. The existing distance measure between average-pooled sentence embeddings (orange) cannot clearly capture the distances of semantically-aligned token pairs (blue).}
\label{fig:concept}
\end{figure}

In this work, we propose an analytical method based on optimal transport to analyze existing approaches that leverage a pretrained model.
We consider a sentence similarity measure a solution to a transportation problem, which aims to transport a collection of contextualized tokens in a sentence to the ones in another sentence.
As byproducts of the problem, we obtain a cost matrix and a transportation matrix, which encode the similarities of all token pairs across sentences and their contributions to the sentence similarity, respectively.
Using this analytical method, we point out that the existing approaches suffer from the rank-1 constraint in the transportation matrix;
this eventually keeps the model from effectively capturing the similarities of semantically-aligned token pairs into sentence similarity.
For example, considering transportation in a contextualized embedding space (Figure \ref{fig:concept}), the distance between averaged token embeddings (orange arrows) cannot clearly represent the distance of semantically-aligned token pairs (blue arrows).

To resolve the above limitation and enhance the interpretability of a model, we present a novel distance measure and a contrastive learning framework that optimizes the distance between sentences.
First, we apply optimal transport in a contextualized embedding space and leverage the optimal solution for a relaxed transportation problem as our distance measure.
This sentence distance is composed of the distances of semantically-aligned token pairs; 
this makes the result easily interpretable.
Furthermore, we present a contrastive learning framework that adopts the proposed distance to finetune the model with token-level supervision.
It optimizes the model to learn the relevance of semantically-aligned token pairs from that of sentence pairs, which further enhances interpretability.

We extensively evaluate our approach and validate the effectiveness of its sentence similarity and interpretation.
The comparison on 7 STS benchmarks supports the superiority of sentence similarity predicted by the model trained by our framework.
In particular, the evaluation on 2 interpretable-STS datasets demonstrates that the proposed distance measure finds out semantically relevant token pairs that are more consistent with human judgement compared to other baseline methods.
Our qualitative analysis shows that both the token alignment and their similarity scores from our model serve as useful resources for end-users to better understand the sentence similarity.

\section{Related work}
\label{sec:related}
\subsection{Semantic textual similarity}
\label{subsec:relatedsts}
Most recent studies tried to leverage a pretrained language model with various model architectures and training objectives for STS tasks, achieving the state-of-the-art performance.
In terms of model architecture, \citet{devlin-etal-2019-bert} focus on exhaustive cross-correlation between sentences by taking a concatenated text of two sentences as an input, while \citet{reimers-gurevych-2019-sentence} improve scalability based on a Siamese network and \citet{Humeau2020Poly-encoders:} adopt a hybrid approach.
Along with the progress of model architectures, many advanced objectives for STS tasks were proposed as well.
Specifically, \citet{reimers-gurevych-2019-sentence} mainly use the classification objective for an NLI dataset,
and \citet{wu2020clear} adopt contrastive learning to utilize self-supervision from a large corpus.
\citet{yan-etal-2021-consert,gao2021simcse} incorporate a parallel corpus such as NLI datasets into their contrastive learning framework.

Despite their effectiveness, the interpretability of the above models for STS tasks was not fully explored \citep{belinkov-glass-2019-analysis}.
One related task is interpretable STS, which aims to predict chunk alignment between two sentences \citep{agirre-etal-2016-semeval-2016}.
For this task, a variety of supervised approaches were proposed based on neural networks \citep{konopik-etal-2016-uwb,lopez-gazpio-etal-2016-iubc}, linear programming \citep{tekumalla-jat-2016-iiscnlp}, and pretrained models \citep{ijcai2020-0333}.
However, these methods cannot predict the similarity between sentences because they focus on finding chunk alignment only.
To the best of our knowledge, no previous approaches based on a pretrained model have taken into account both sentence similarity and interpretation.

\subsection{Optimal transport}
\label{subsec:ot}
Optimal transport~\citep{monge1781memoire} has been successfully applied to many applications in natural language processing~\citep{li-etal-2020-improving-text,xu-etal-2021-vocabulary}, by the help of its ability to find a plausible correspondence between two objects~\citep{9679111,LeeLY21}.
For example, \citet{pmlr-v37-kusnerb15} adopt optimal transport to measure the distance between two documents with pretrained word vectors.
\citet{zhao-etal-2019-moverscore} adopt optimal transport for evaluating text generation and \citet{Zhang*2020BERTScore:} take a greedy approach leveraging pretrained language model.
In addition, \citet{swanson-etal-2020-rationalizing} discover the rationale in text-matching via optimal transport, thereby improving model interpretability.

One well-known limitation of optimal transport is that finding the optimal solution is computationally intensive, and thus approximation schemes for this problem have been extensively researched~\citep{grauman2004fast,shirdhonkar2008approximate}.
To get the solution efficiently, \citet{NIPS2013_af21d0c9} provides a regularizer inspired by a probabilistic theory and then uses Sinkhorn's algorithm.
\citet{pmlr-v37-kusnerb15} relax the problem to get the quadratic-time solution by removing one of the constraints, and \citet{wu-etal-2018-word} introduce a kernel method to approximate the optimal transport.


\section{Method}
\label{sec:method}
We first analyze the similarity measure used by existing models from the perspective of a transportation problem.
Considering the above analysis, we present a novel distance measure and a contrastive sentence learning framework to enhance the interpretability of a sentence similarity model.

\subsection{Distance as a transportation problem}
\label{subsec:otanalysis}
We briefly explain the transportation problem and how to interpret the total transportation cost as a distance measure.
A transportation problem consists of three components: states before and after transportation, and a cost matrix.
In general, the two states are represented in high-dimensional simplex, i.e., $\mathbf{d}^1 \in \Sigma^{d_1}$ and $\mathbf{d}^2 \in \Sigma^{d_2}$, where each dimension implies a specific location with a non-negative quantity.
The cost matrix $\mathbf{M} \in \mathbb{R}^{d_1 \times d_2}$ encodes the unit transportation cost from location $i$ to $j$ into $\mathbf{M}_{i,j}$.
In this situation, we search the transportation plan to transport from $\mathbf{d}^1$ to $\mathbf{d}^2$ with the minimum cost.
Using the above notations, the optimization problem is written as follows:
\begin{gather}
\label{eq:otproblem}
    \underset{\mathbf{T}\in\mathbb{R}^{d_1\times d_2}_{\ge0}}{\mathrm{minimize}} \sum_{i,j} \mathbf{T}_{i,j} \mathbf{M}_{i,j} \\
    \text{ subject to } \mathbf{T}^\top\Vec{\mathbf{1}} = \mathbf{d}^2\text{, }\mathbf{T} \Vec{\mathbf{1}} = \mathbf{d}^1,\nonumber
\end{gather}
where each entry of the transportation matrix $\mathbf{T}_{i,j}$ indicates how much quantity is transferred from location $i$ to $j$.
The optimal solution to this problem is called optimal transport, which is also known as earth mover's distance (EMD):
\begin{equation}
    d_{\mathbf{M}}^{\mathrm{EMD}} := \sum_{i,j} \mathbf{T}^*_{i,j} \mathbf{M}_{i,j} \label{eq:emd}.
\end{equation}

In Equation \eqref{eq:emd}, the distance is computed by the sum of element-wise multiplications of the optimal transportation matrix $\mathbf{T}^*$ and the cost matrix $\mathbf{M}$.
In this sense, EMD considers the optimality of distance when combining unit costs in $\mathbf{M}$.
That is, the priority of each unit cost when being fused to the distance is encoded in the transportation matrix, which serves as a useful resource for analyzing the distance.

\subsubsection{Example: Average pooling}
\label{subsubsec:ot4avgpool}
We express cosine similarity with average pooling as a transportation problem and analyze its properties in terms of the transportation matrix.
Note that this similarity measure is widely adopted in most of the previous studies~\cite{reimers-gurevych-2019-sentence,wu2020clear,gao2021simcse}.
Formally, for a sentence of length $L$, the sentence embedding is generated by applying average pooling to $L$ contextualized token embeddings, i.e., $\mathbf{s} = \frac{1}{L} \sum_{i=1}^L \mathbf{x}_i$, where $\mathbf{x}_i$ is the $i$-th token embedding obtained from a pretrained model.
Using the sentence embeddings, the sentence similarity is defined by 
\begin{equation}
    s^{\text{AVG}} = \cos (\mathbf{s}^1, \mathbf{s}^2) = \frac{\mathbf{s}^{1}{}^\top \mathbf{s}^2}{\|\mathbf{s}^1\|\|\mathbf{s}^2\|}. \nonumber
\end{equation}
This average pooling-based sentence similarity can be converted into the distance, $d^{\text{AVG}}=1-s^{\text{AVG}}$, described by the token embeddings as follows:
\begin{equation}
\label{eq:avg_dist}
d^{\text{AVG}} = 1- \sum_{i=1}^{L_1}\sum_{j=1}^{L_2} \frac{1}{L_1L_2}  \frac{\|\mathbf{x}_i^1\|\|\mathbf{x}_j^2\|}{\left\|\mathbf{s}^1\right\|\left\|\mathbf{s}^2\right\|}  \frac{\mathbf{x}_i^1{}^\top \mathbf{x}_j^2}{\|\mathbf{x}_i^1\|\|\mathbf{x}_j^2\|}. \nonumber
\end{equation}
From the perspective of Equation~\eqref{eq:otproblem}, this distance is interpreted as a naive solution of a special transportation problem,
where the cost matrix and the transportation matrix are
\begin{align}
    \mathbf{M}_{i,j}^\mathrm{AVG} &= \frac{\|\mathbf{s}^1\|\|\mathbf{s}^2\|}{\|\mathbf{x}_i^1\|\|\mathbf{x}_j^2\|} - \cos ( \mathbf{x}_{i}^1, \mathbf{x}_{j}^{2} ), \nonumber \\
    \mathbf{T}_{i,j}^{\mathrm{AVG}} &= \frac{1}{L_1L_2} \frac{\|\mathbf{x}_i^1\|\|\mathbf{x}_j^2\|}{\left\|\mathbf{s}^1\right\|\left\|\mathbf{s}^2\right\|}. \label{eq:avg_sol}
\end{align}

Each entry of the cost matrix includes negative cosine similarities between token embeddings, and the contribution of each token pair to the sentence distance (i.e., the transportation matrix) is determined by the norms of the token embeddings.
In theory, the rank of the transportation matrix is constrained to be one, which prevents effective integration of the token distances into the sentence distance.
In practice, it is impossible to involve only semantically-aligned token pairs across sentences because all possible token pairs are considered by the products of their norms.
From this analysis, we point out that the average pooling-based similarity is not effective enough to capture the token correspondence between sentences.

\subsection{Relaxed optimal transport distance for contextualized token embeddings}
\label{subsec:rcmd}
To resolve the ineffectiveness of the existing measure, we introduce a novel distance measure based on optimal transport.
We first define a transportation problem that considers semantic relevance in a contextualized embedding space.
Given the token embeddings of two sentences from a pretrained language model, we construct a cost matrix $\mathbf{M}^{\mathrm{CMD}} \in \mathbb{R}^{L_1 \times L_2}$ that encodes token similarities using cosine distance, and define the state vectors for the two sentences as one vectors normalized by their sentence lengths $\mathbf{d}^1:= \frac{1}{L_1}\Vec{\mathbf{1}}$ and $\mathbf{d}^2 := \frac{1}{L_2}\Vec{\mathbf{1}}$.
As discussed in Section~\ref{subsec:otanalysis}, we consider the optimal solution to this problem as a distance measure named contextualized token mover's distance (CMD):
\begin{gather}
\label{eq:rcmd_cost}
    \mathbf{M}_{i,j}^\mathrm{{CMD}} := 1 - \cos\left(\mathbf{x}_i^1, \mathbf{x}_j^2\right), \nonumber\\
    d_{\mathbf{M}}^{\mathrm{CMD}} := \sum_{i,j} \mathbf{T}^*_{i,j} \mathbf{M}_{i,j}^{\mathrm{CMD}}. \nonumber
\end{gather}

However, finding $\mathbf{T}^*$ incurs huge computational complexity of $O(L^3\log L)$ where $L=\max (L_1, L_2)$ \citep{villani2008optimal}.
For this reason, we relax the optimization problem by removing the first constraint, $\mathbf{T}^\top\Vec{\mathbf{1}} = \mathbf{d}'$, similar to \citet{pmlr-v37-kusnerb15}.
The optimal solution for this relaxed transportation problem is found in $O(L^2)$, keeping the rank of the transportation matrix larger than one.
In the end, the optimal transportation matrix and the corresponding distance named relaxed CMD (RCMD) are derived as follows:
\begin{gather}
    \resizebox{0.86\linewidth}{!}{%
    $\displaystyle
    \mathbf{T}^{\mathrm{RCMD_1}}_{i,j} = 
    \begin{cases}
    \frac{1}{L_1} & \text{if } j = \argmin_{j'} \mathbf{M}_{i,j'}^{\mathrm{CMD}}\\
    0 & \text{otherwise,}
    \end{cases}
    $} \nonumber \\ 
    d_{\mathbf{M}}^{\mathrm{RCMD_1}} := \frac{1}{L_1}\ {\sum_i \min_j \mathbf{M}_{i, j}^\mathrm{CMD}}.
\label{eq:rcmd_sol}
\end{gather}
Similarly, the elimination of the second constraint, $\mathbf{T} \Vec{\mathbf{1}} = \mathbf{d}$, results in $\mathbf{T}^{\mathrm{RCMD_2}}$ and $d_{\mathbf{M}}^{\mathrm{RCMD_2}}$, where the solutions for the two relaxed problems use $\min$ operation on the cost matrix in a row-wise and a column-wise manner, respectively.
Note that $\mathbf{T}^{\mathrm{RCMD_1}}$ represents the token-level binary alignment from the first sentence to the second sentence and accordingly the final distance is computed by averaging all the distances of the aligned token pairs.
Also, it is obvious that $\mathbf{T}^{\mathrm{RCMD_1}}$ has a much higher rank than $\mathbf{T}^{\mathrm{AVG}}$, which implies that it can express more complex token-level semantic relationship between two sentences.

We remark that our solution provides better interpretability of semantic textual similarity compared to the case of average pooling.
For the sentence distance in Equation~\eqref{eq:avg_sol}, $\mathbf{T}^{\mathrm{AVG}}$ assigns non-zero values to all token pairs that include irrelevant pairs;
this makes it difficult to interpret the result.
On the contrary, $\mathbf{T^{\mathrm{RCMD_1}}}$ in Equation~\eqref{eq:rcmd_sol} is designed to explicitly involve the most relevant token pairs across sentences for the sentence distance, which allows us to interpret the result easily. 

\subsection{Contrastive sentence similarity learning with semantically-aligned token pairs}
\label{subsec:cl-rcmd}
We present a contrastive learning framework for \proposeddist (\proposed) that incorporates \proposeddist into the state-of-the-art contrastive learning framework.
To this end, we convert \proposeddist to the corresponding similarity by $s_{\mathbf{M}}^{\mathrm{RCMD_1}} = 1-d_{\mathbf{M}}^{\mathrm{RCMD_1}}$:
\begin{gather}
    s_{\mathbf{M}}^{\mathrm{RCMD_1}}(\mathbf{s}^1,\mathbf{s}^2) = \frac{1}{L_1} \ {\sum_{i=1}^{L_1} \max_j \cos(\mathbf{x}_i^1, \mathbf{x}_{j}^2)} \nonumber.
\end{gather}
$s_{\mathbf{M}}^\mathrm{RCMD_2}$ is computed in the same manner as well, and we average them to consider bidirectional semantic alignment between two sentences;
this provides diverse gradient signals during optimization.
The final similarity is described by
\begin{gather}
    \resizebox{0.95\linewidth}{!}{%
    $\displaystyle s_{\mathbf{M}}^{\mathrm{RCMD}}(\mathbf{s}^1,\mathbf{s}^2) := \frac{1}{2}\left(s_{\mathbf{M}}^{\mathrm{RCMD_1}}(\mathbf{s}^1,\mathbf{s}^2) + s_{\mathbf{M}}^{\mathrm{RCMD_2}}(\mathbf{s}^1,\mathbf{s}^2)\right) \nonumber$%
    }.
\end{gather}
Adopting this similarity measure, the contrastive learning objective for the $i$-th sentence pair in a training batch is defined as follows:
\begin{gather}
\label{eq:cl-rcmd}
    \resizebox{0.95\linewidth}{!}{%
    $\displaystyle -\log\frac{\exp(s_{\mathbf{M}}^{\mathrm{RCMD}}(\mathbf{s}^i,\mathbf{s}_{+}^i)/\tau)}{\sum_{j=1}^B(\exp(s_{\mathbf{M}}^{\mathrm{RCMD}}(\mathbf{s}^i,\mathbf{s}_{+}^j)/\tau)+\exp(s_{\mathbf{M}}^{\mathrm{RCMD}}(\mathbf{s}^i,\mathbf{s}_{-}^j)/\tau))}, \nonumber$%
    }
\end{gather}
where $\tau$ is the temperature parameter and $B$ is the batch size.
Following~\citep{gao2021simcse}, \proposed uses the other sentences in the batch to generate negative pairs.
\begin{table*}
\begin{tabular}{@{}ccccccccc@{}}
\toprule
Model & STS12 & STS13 & STS14 & STS15 & STS16 & STS-B & SICK-R & Avg \\ \midrule
\bertbase-avg & 29.12 & 59.96 & 47.22 & 60.61 & 63.72 & 47.20 & 58.25 & 52.30 \\
S\bertbase{}\textsuperscript{$\dagger$} & 70.97 & 76.53 & 73.19 & 79.09 & 74.30 & 77.03 & 72.91 & 74.86 \\
S\bertbase-flow\textsuperscript{$\dagger$} & 69.78 & 77.27 & 74.35 & 82.01 & 77.46 & 79.12 & 76.21 & 76.60 \\
S\bertbase-whitening\textsuperscript{$\dagger$} & 69.65 & 77.57 & 74.66 & 82.27 & 78.39 & 79.52 & 76.91 & 77.00 \\
\simcsecls-\bertbase{}\textsuperscript{$\dagger$} & 75.30 & 84.67 & 80.19 & 85.40 & 80.82 & 84.25 & 80.39 & 81.57 \\
\simcseavg-\bertbase & \textbf{75.88} & 83.28 & 80.26 & 86.06 & 81.33 & 84.91 & 79.94 & 81.67 \\
\proposed-\bertbase & 75.23 & \textbf{85.06} & \textbf{80.99} & \textbf{86.26} & \textbf{81.50} & \textbf{85.21} & \textbf{80.49} & \textbf{82.11} \\ \midrule
\robertabase-avg & 32.50 & 55.78 & 45.00 & 60.61 & 61.68 & 55.31 & 61.66 & 53.22 \\
S\robertabase{}\textsuperscript{$\dagger$} & 71.54 & 72.49 & 70.80 & 78.74 & 73.69 & 77.77 & 74.46 & 74.21 \\
S\robertabase-whitening\textsuperscript{$\dagger$} & 70.46 & 77.07 & 74.46 & 81.64 & 76.43 & 79.49 & 76.65 & 76.60 \\
\simcsecls-\robertabase{}\textsuperscript{$\dagger$} & \textbf{76.53} & 85.21 & \textbf{80.95} & 86.03 & 82.57 & 85.83 & 80.50 & 82.52 \\
\simcseavg-\robertabase & 75.75 & 85.10 & 80.85 & 85.95 & 83.33 & 85.55 & 79.41 & 82.28 \\
\proposed-\robertabase & 75.68 & \textbf{85.76} & 80.92 & \textbf{86.58} & \textbf{83.48} & \textbf{85.89} & \textbf{81.01} & \textbf{82.76} \\ \bottomrule
\end{tabular}
\caption{\label{tab:stsbenchmark}
The results on 7 STS benchmarks.
We measure Spearman correlation on all the examples \citep{gao2021simcse}.
$\dagger$ indicates the baseline results reported in their original papers.
}
\end{table*}

We argue that \proposed enhances both the sentence similarity and its interpretability in the following aspects.
First, \proposed alleviates the catastrophic forgetting of pretrained semantics during the finetuning process.
Its token-level supervision is produced by leveraging the textual semantics encoded in a pretrained checkpoint, because token pairs are semantically aligned according to their similarities in the contextualized embedding space.
Namely, \proposed updates the parameters to improve the quality of sentence similarity while less breaking token-level semantics in the pretrained checkpoint.
Furthermore, \proposed directly distills the relevance of a sentence pair into the relevance of semantically-aligned token pairs.
In this sense, our contextualized embedding space effectively captures the token-level semantic relevance from training sentence pairs, which provides better interpretation for its sentence similarity.
%

%

\section{Experiments}
\label{sec:experiments}
To analyze our approach in various viewpoints, we design and conduct experiments that focus on the following three research questions:
\begin{itemize}
    \setlength\itemsep{-0.3em}
    \item \textbf{RQ1} Does \proposed effectively measure sentence similarities using a pretrained language model?
    \item \textbf{RQ2} Does \proposed provide the interpretation of sentence similarity which is well aligned with human judgements?
    \item \textbf{RQ3} Does \proposed efficiently compute its sentence similarity for training and inference?   
\end{itemize}

\subsection{Training details}
\label{subsec:training}
We finetune a pretrained model using \proposed in the following settings.
Following previous work \citep{gao2021simcse}, we use NLI datasets with hard negatives:
SNLI~\citep{bowman-etal-2015-large} and MNLI~\citep{williams-etal-2018-broad}.
We use a pretrained backbone attached with a single head, which is the same with \citep{gao2021simcse}.
As the initial checkpoint of the pretrained models, we employ \texttt{bert-base-uncased} and \texttt{roberta-base} provided by huggingface \citep{devlin-etal-2019-bert,liu2019roberta}.
Adam optimizer is used with the initial learning rate $5e-5$ and linear decay schedule.
Fp16 training is enabled where the maximum batch size is 128 on a single V100 GPU, and the softmax temperature is set to $\tau=0.05$ \citep{gao2021simcse}.
The training is proceeded with 4 different random seeds and the best model is chosen using the best Spearman correlation on STSb validation set which is evaluated every 250 steps during training.

\subsection{Semantic textual similarity}
\label{subsec:sts}
We evaluate the similarity model finetuned by \proposed for STS task to quantitatively measure the quality of sentence similarity (\textbf{RQ1}).

\paragraph{Metric}
We measure Spearman correlation for each of seven STS benchmarks and calculate their average to compare the capability of representing sentences in general \citep{conneau-kiela-2018-senteval}.

\paragraph{Baselines}
We select the baselines that leverage a pretrained model, and they turn out to outperform other traditional baselines.
We only list the baseline names for \bertbase below;
the names for \robertabase are obtained by replacing \bertbase with \robertabase.
\begin{itemize}
    \setlength\itemsep{-0.3em}
    \item \textbf{\bertbase-avg} generates sentence embeddings by averaging the token embeddings from \bertbase without finetuning.
    It indicates zero-shot performance of a checkpoint.
    
    \item \textbf{S\bertbase} \citep{reimers-gurevych-2019-sentence} is a pioneering work to finetune a pretrained model for sentence embeddings. 
    It trains a Siamese network using NLI datasets. 
    
    \item \textbf{\simcsecls-\bertbase}~\citep{gao2021simcse} adopts a contrastive learning framework \citep{pmlr-v119-chen20j} using SNLI and MNLI datasets. The contextualized embedding of \texttt{[CLS]} is used as a sentence embedding.
    
    \item \textbf{\simcseavg-\bertbase}~\citep{gao2021simcse} is the same with \simcsecls-\bertbase except that it performs average pooling on token embeddings to obtain a sentence embedding.
\end{itemize}

\paragraph{Result}
Table \ref{tab:stsbenchmark} reports Spearman correlation for each dataset and their average.
For most of the datasets, \proposed shows higher correlation compared to the state-of-the-art baselines.
In particular, for STS14, STS15, SICK-R datasets, \proposed-\bertbase achieves comparable performance to \simcsecls-\robertabase whose backbone language model is pretrained with 10 times larger data compared to \bertbase.
This implies that finetuning with token-level supervision from \proposed achieves the performance as good as using an expensively pretrained checkpoint.

\subsection{Interpretable semantic textual similarity}
\label{subsec:ists}
Next, we measure the performance of our approach on interpretable STS (iSTS) tasks in order to validate that \proposed embeds a sufficient level of interpretability even without any supervision (i.e., labeled training data) about semantically-aligned chunk pairs (\textbf{RQ2}).

\paragraph{Experimental setup}
We utilize the ``images'' and ``headlines'' data sources included in SemEval2016 Task 2: iSTS \citep{agirre-etal-2016-semeval-2016}.
We measure the agreement between human judgement (gold semantic alignment across sentences) and the contributions of all token pairs to sentence similarity (element-wise multiplication of $(1-\mathbf{M})$ and $\mathbf{T}$).
One challenge to use our similarity model for this task is to convert token pair contributions into chunk-level alignment.
First, we summarize token pair contributions into chunk pair contributions by applying simple average pooling based on the chunk mapping represented by $c(i) = \{ k | \text{is\_overlap} (c_i, t_k) \}$, where $c_i$ is the $i$-th chunk and $t_k$ is the $k$-th token in a sentence.\footnote{We use gold standard chunking information to focus on alignment only, which is the second subtrack in iSTS.}
Then, to obtain the alignment based on the pairwise chunk contributions, we design a criterion for selecting confident chunk pairs $(i, j)$ as follows:
\begin{gather}
    \mathbf{C}_{i,j} = \frac{1}{|c_1(i)||c_2(j)|} \sum_k^{c_1(i)} \sum_l^{c_2(j)} \mathbf{T}_{k,l} \mathbf{M}_{k,l}, \nonumber\\
    \resizebox{\linewidth}{!}{%
    $\displaystyle
    a(i, j) = \mathbb{I}[j = \argmax_{j'}\mathbf{C}_{i,j'}]
    \cdot\mathbb{I}[i = \argmax_{i'} \mathbf{C}_{i',j}]. \nonumber
    $}
\end{gather}
Using the aligned chunk pairs obtained by each method, we compute the alignment F1 score as the evaluation metric, which indicates the agreement between human judgement and chunk contribution.\footnote{We employ alignment F1 score implemented in the evaluation script provided by the task organizer.}
We consider eight different configurations to investigate the effectiveness of the following components: 1) sentence similarity, 2) contrastive learning, and 3) pretrained checkpoints.
\begin{table}
\centering
\begin{tabular}{@{}cccccc@{}}
\toprule
Model & images & headlines \\ \midrule
\bertbase-avg & 82.45 & 85.98 \\
\bertbase-\proposeddist & \textbf{83.00} & \textbf{88.25} \\ \midrule
\simcseavg-\bertbase & 82.98 & 85.80 \\
\proposed-\bertbase & \textbf{87.25} & \textbf{90.55} \\ \midrule
\robertabase-avg & 61.68 & 52.01 \\
\robertabase-\proposeddist & \textbf{82.44} & \textbf{88.92} \\ \midrule
\simcseavg-\robertabase & 73.66 & 77.30 \\
\proposed-\robertabase & \textbf{84.93} & \textbf{88.45} \\ \bottomrule
\end{tabular}
\caption{The results on SemEval2016 task 2: iSTS.}
\label{tab:ists}
\end{table}

\begin{figure*}[t]
\begin{subfigure}[b]{.33\textwidth}
  \centering
  \includegraphics[width=\linewidth]{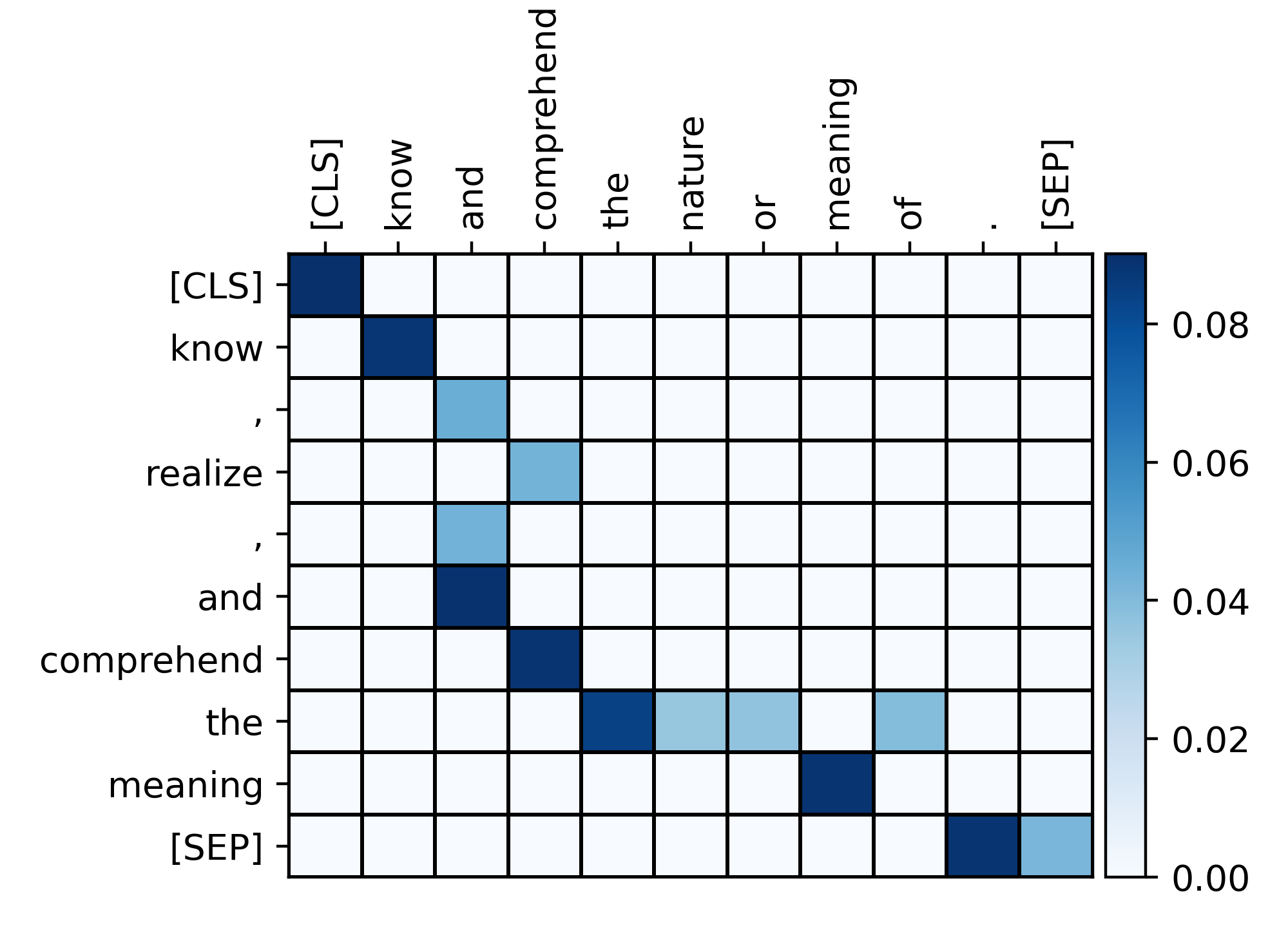}
  \caption{Positive, \proposed}
  \label{fig:token-heatmap-proposed-pos}
\end{subfigure}
\begin{subfigure}[b]{.33\textwidth}
  \centering
  \includegraphics[width=0.9\linewidth]{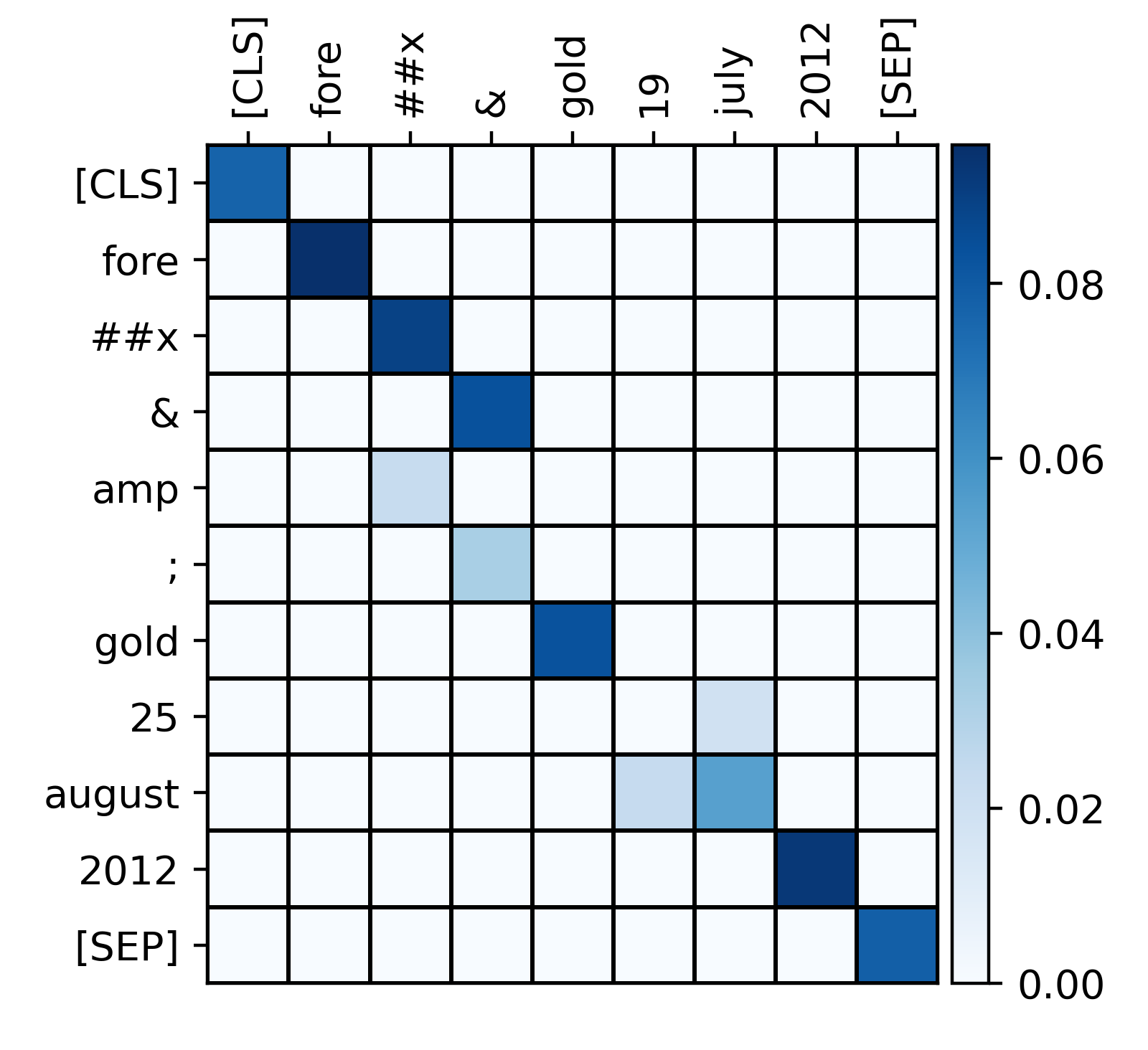}
  \caption{Neutral, \proposed}
  \label{fig:token-heatmap-proposed-neu}
\end{subfigure}
\begin{subfigure}[b]{.33\textwidth}
  \centering
  \includegraphics[width=\linewidth]{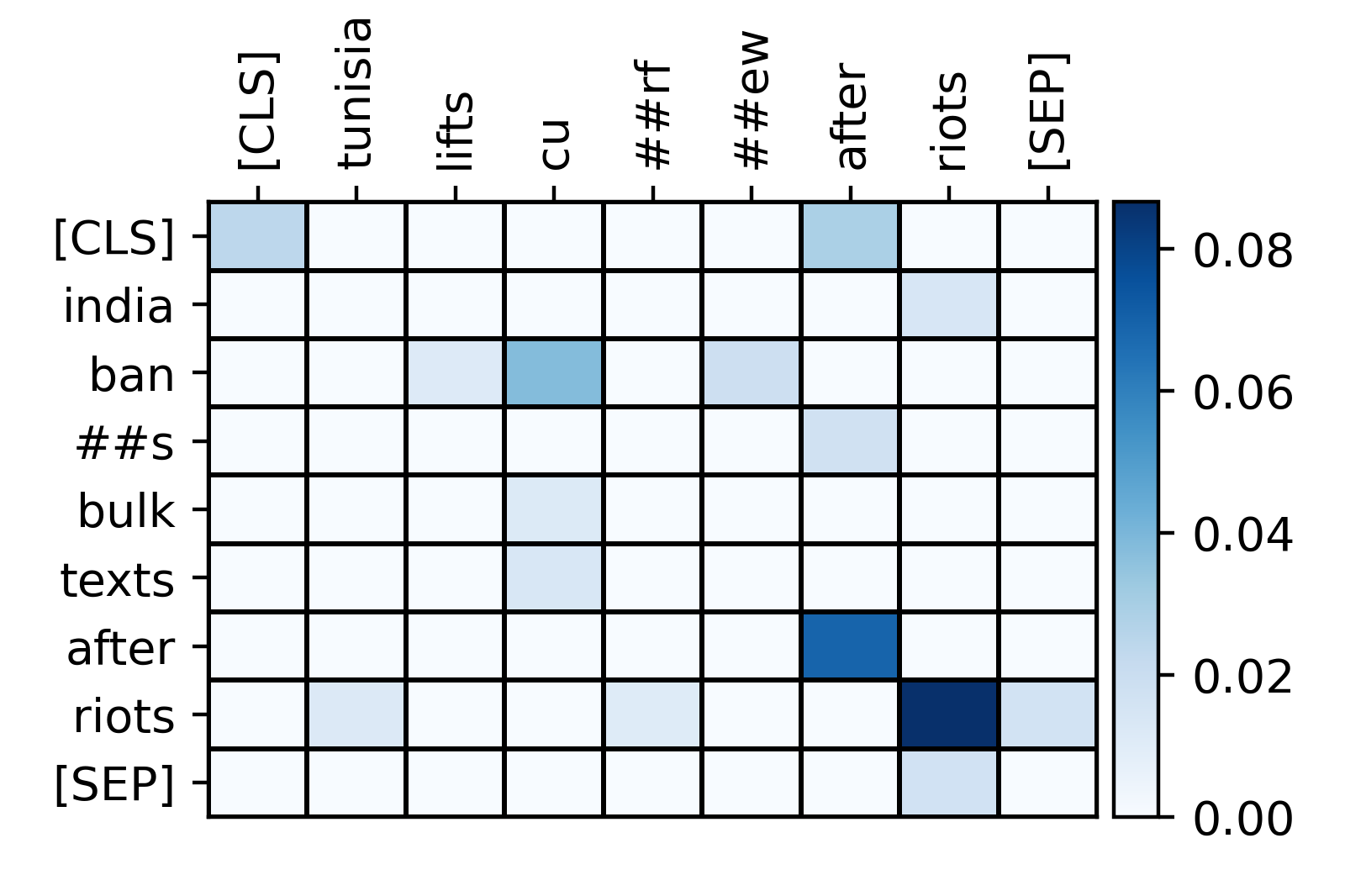}
  \caption{Negative, \proposed}
  \label{fig:token-heatmap-proposed-neg}
\end{subfigure}
\begin{subfigure}[b]{.33\textwidth}
  \centering
  \includegraphics[width=\linewidth]{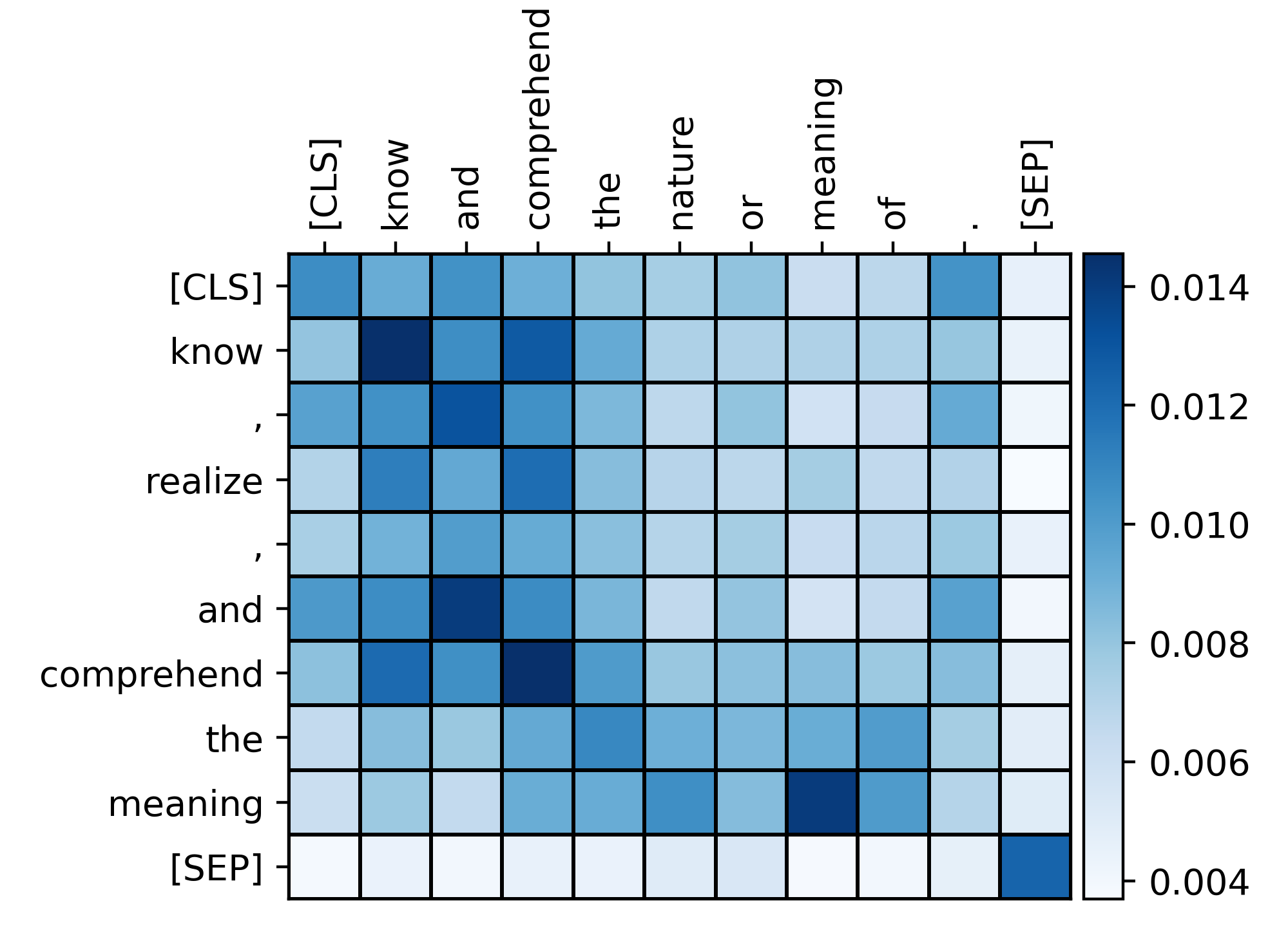}
  \caption{Positive, \simcseavg}
  \label{fig:token-heatmap-avg-pos}
\end{subfigure}%
\begin{subfigure}[b]{.33\textwidth}
  \centering
  \includegraphics[width=0.9\linewidth]{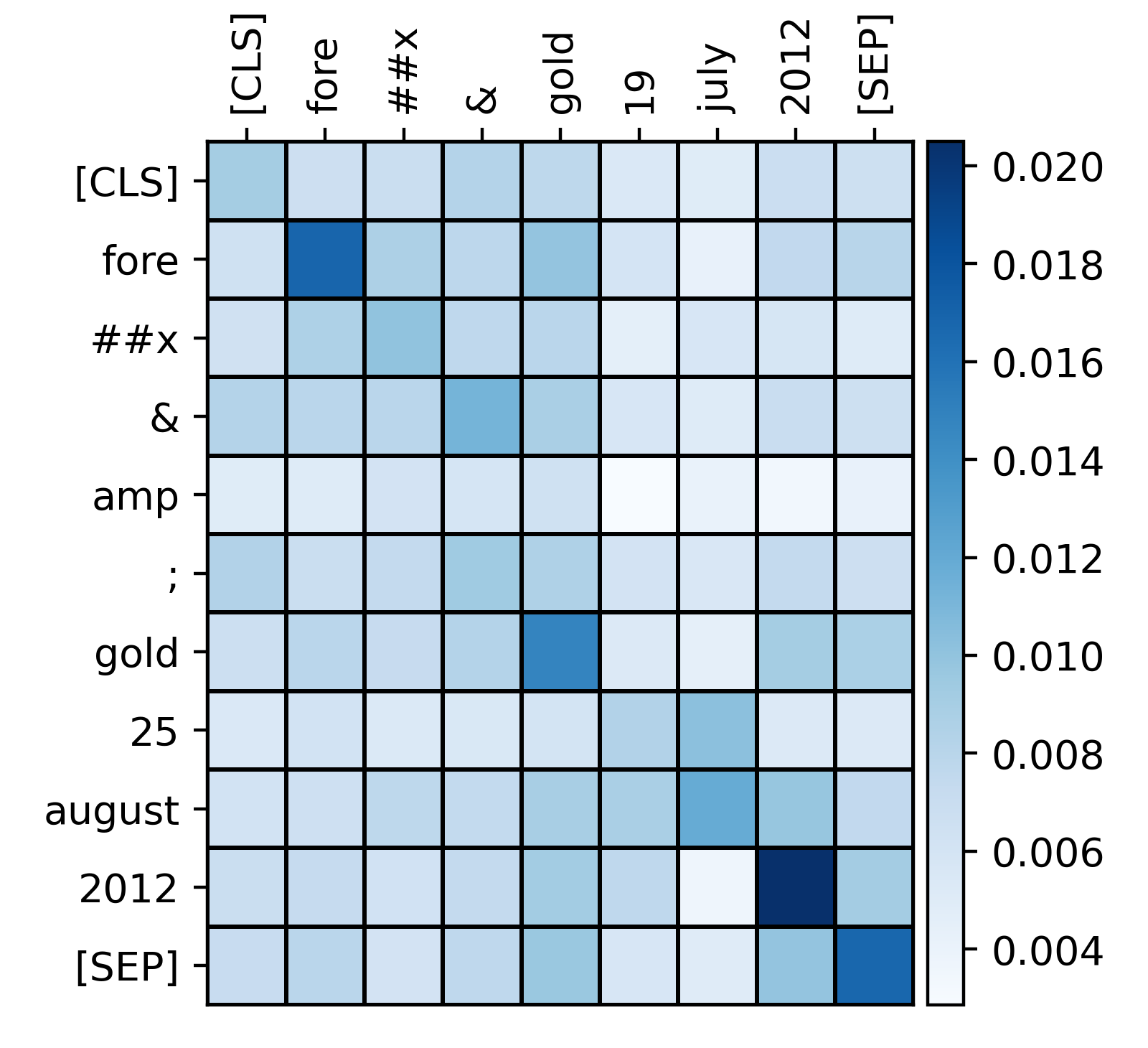}
  \caption{Neutral, \simcseavg}
  \label{fig:token-heatmap-avg-neu}
\end{subfigure}
\begin{subfigure}[b]{.33\textwidth}
  \centering
  \includegraphics[width=\linewidth]{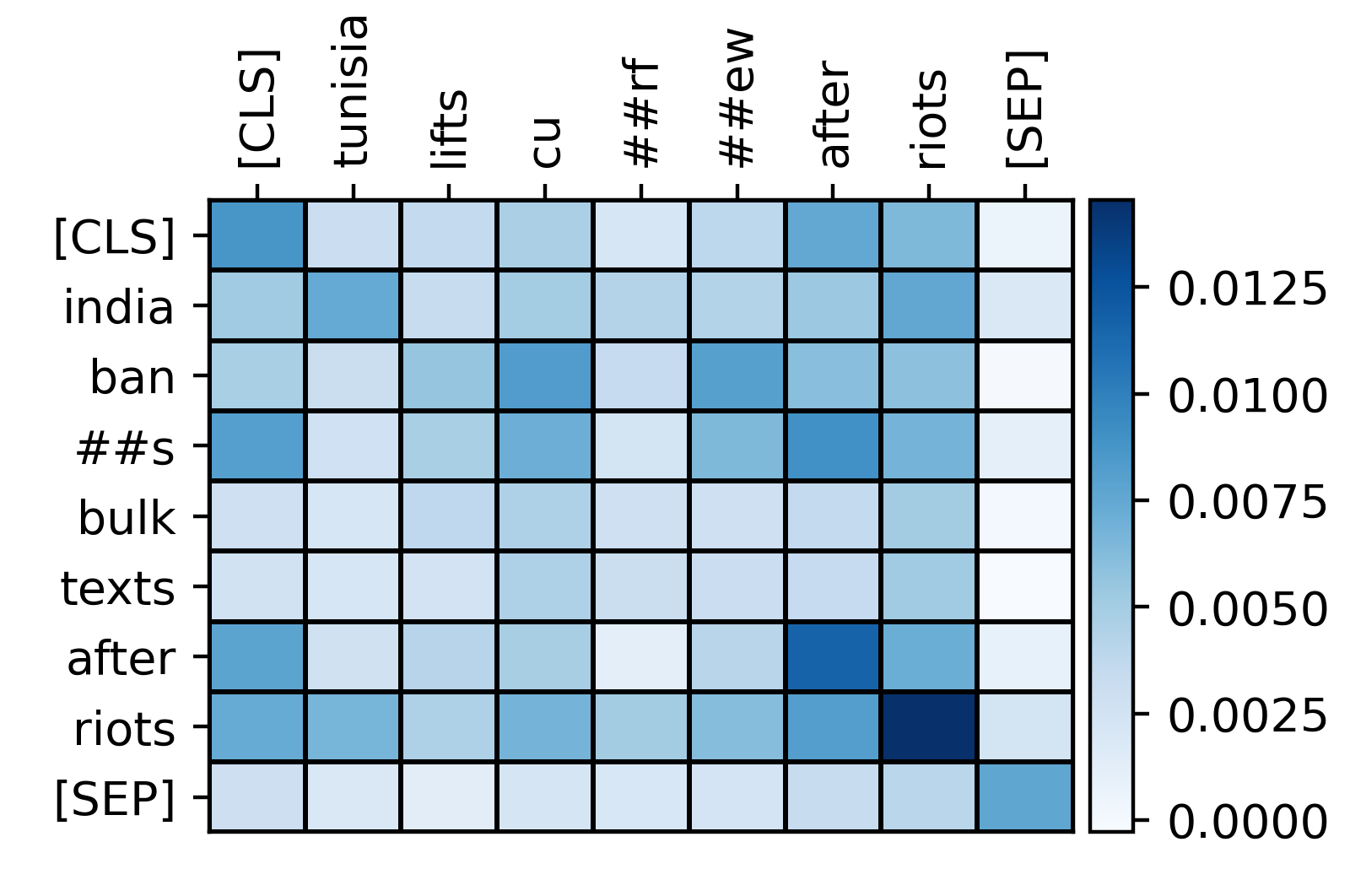}
  \caption{Negative, \simcseavg}
  \label{fig:token-heatmap-avg-neg}
\end{subfigure}
\caption{Token pair contribution heatmaps. We use the model finetuned from \bertbase for this experiment.}
\label{fig:token-heatmap}
\end{figure*}

\paragraph{Result}
Table~\ref{tab:ists} shows the clear tendency of iSTS performance with respect to each of the above components.
First of all, the token pair contribution from \proposeddist is more consistent with human judgement than that from average pooling.
\proposeddist improves alignment F1 scores even without finetuning (\bertbase-\proposeddist and \robertabase-\proposeddist), indicating that \proposeddist effectively discovers the token-level relevance encoded inside a pretrained checkpoint.
In addition, the alignment F1 score increases when we finetune a model using \proposed.
Notably, \proposed-\bertbase successfully improves the alignment F1 score whereas \simcseavg-\bertbase does not.
This result shows that finetuning a model using the similarity measure based on semantically-aligned token pairs (i.e., fine-grained supervision induced by \proposeddist) further enhances the interpretability of a model.

\subsection{Qualitative analysis}
\label{subsec:qualanal}
We qualitatively analyze the sentence similarity from the perspective of the transportation problem in order to demonstrate that a model trained by \proposed provides clear and accurate explanation (\textbf{RQ2}).
To this end, we visualize the contribution of token pairs obtained from \proposed-\bertbase and that from \simcseavg-\bertbase, and then clarify how their sentence similarity is computed differently from each other.
Three sentence pairs are randomly selected from STS13 dataset.
Figure \ref{fig:token-heatmap} illustrates the token pair contribution heatmap for positive, neutral, and negative sentence pairs.

\paragraph{\proposed vs. \simcseavg}
Overall, \proposed aligns two sentences better than the baseline.
To be specific, \proposed effectively highlights the contributions of semantically-relevant token pairs and excludes the other contributions (Figure \ref{fig:token-heatmap} upper).
On the contrary, \simcseavg fails to represent meaningful token-level relevance for sentence similarity (Figure \ref{fig:token-heatmap} lower).
The rank-1 constraint of \simcseavg prevents the model from getting any plausible alignment between two sentences, while it simply tunes the contributions of all possible token pairs at once.
We emphasize that the superfluous correlation in the heatmap not only inhibits the capability to capture sentence similarity, but also makes it difficult for humans to understand how sentence similarity is computed.

\paragraph{Case study on positive, neutral, and negative sentence pairs}
For the positive pair (Figure \ref{fig:token-heatmap} left), \proposed clearly matches all semantically-aligned token pairs including the linking words (\{``,'', ``and''\}--\{``and''\}), synonyms (\{``realize'', ``comprehend''\}--\{``comprehend''\}), and omitted contexts (\{``the''\}--\{``the nature of'', ``of''\}).
For the neutral pair (Figure \ref{fig:token-heatmap} middle), the two sentences have the same lexical structure except for the date.
In this case, \proposed assigns low contributions to the token pairs about day and month (\{``25'', ``august''\}--\{``19'', ``july''\}), while keeping the contributions high for all the other pairs of identical tokens.
Therefore, end-users can clearly figure out which part is semantically different based on their contributions as well as alignment.
In case of the negative pair (Figure \ref{fig:token-heatmap} right), both the models are not able to find any plausible alignment;
\proposed lowers contributions for most of the token pairs except the token pair with identical contents (``after riots'').
That is, end-users also can interpret the negative pair based on the heatmap where semantic correspondence between two sentences does not clearly exist but few overlapped tokens highly contribute to the sentence similarity.

\subsection{Resource evaluation}
\label{subsec:resourceeval}
We measure GPU memory usage and inference time of \proposed-\bertbase to demonstrate that \proposed can be executed on a single GPU and an inference of our model takes almost the same cost to that of the baseline (\textbf{RQ3}).

\begin{table}[]
\centering
\begin{tabular}{@{}ccccc@{}}
\toprule
Batch size & 16 & 32 & 64 & 128 \\ \midrule
\densedist & 7.5 & 22.2 & OOM & OOM \\
\sparsedist & 4.6 & 6.1 & 10.6 & 25.8 \\ \bottomrule
\end{tabular}
\caption{GPU memory usage (GB) of \proposed with various batch sizes. OOM: out-of-memory.}
\label{tab:gpuusage}
\end{table}
\subsubsection{GPU memory usage analysis}
\paragraph{Implementation of \proposeddist}
We implement two variants of \proposeddist, \densedist and \sparsedist, to investigate the effect of exploiting the sparseness in \proposeddist.
Both of them calculate sentence distance by the sum of element-wise multiplications of the cost matrix and the transportation matrix.
For an input sentence pair,
\densedist maintains the full pairwise token distances ($\mathbf{M}^{\mathrm{CMD}}$), 
whereas \sparsedist only keeps the token distances at which the transportation matrix has nonzero values ($\{\mathbf{M}^{\mathrm{CMD}}_{i,j} | \mathbf{T}^{\mathrm{CMD}}_{i,j}\neq 0\}$).
Note that the number of nonzero entries in the transportation matrix of \proposeddist is at most $2L$, which is an order of magnitude smaller than the number of all entries, $L^2$. 

\paragraph{Result}
Table \ref{tab:gpuusage} reports the GPU memory usage during the finetuning process.
For batch-wise contrastive learning, GPU memory requirement becomes $O(B^2)$ in terms of the batch size $B$, because all pairwise sentence similarities within a batch need to be computed.
In this situation, \densedist using a dense matrix drastically increases GPU memory usage  by $O(B^2L^2)$, and as a result, the batch size cannot grow larger than 32.
In contrast, \sparsedist successfully enlarges the batch size up to 128 by exploiting sparseness in the transportation matrix of \proposeddist, which eventually reduces the space complexity to $O(B^2L)$.

\subsubsection{Inference time analysis}
\paragraph{Experimental setup}
We measure the time for predicting the similarities of 512 sentence pairs on a single V100 GPU while increasing the sequence length from 8 to 128, which is the most influential factor for inference time.
We repeat this process 10 times and report the average values.

\begin{figure}
    \centering
    \includegraphics[width=\linewidth]{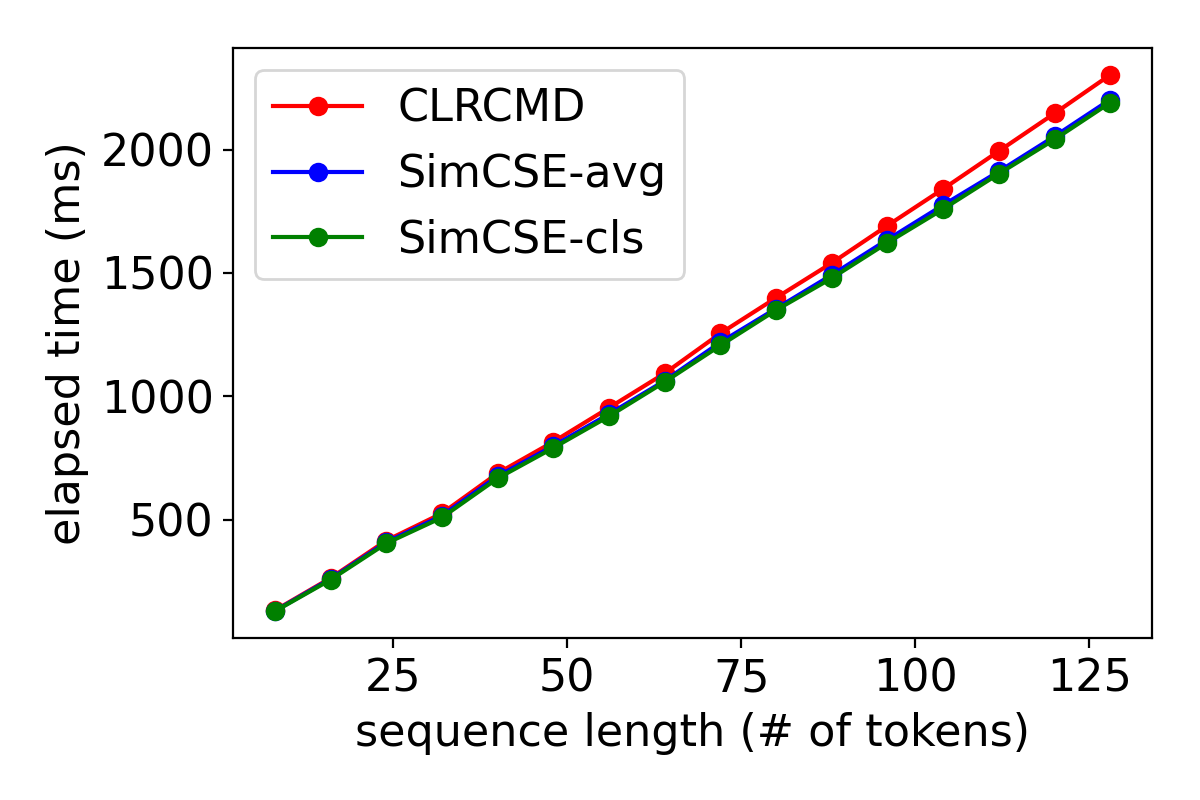}
    \caption{Elapsed time (ms) for the inference of 512 sentence pairs. The result of \simcseavg and \simcsecls are overlapped.}
    \label{fig:inference}
\end{figure}

\paragraph{Result}
Figure \ref{fig:inference} shows the average elapsed time for inference.
The model with \proposeddist takes almost the same inference time as the model with the simple average pooling-based similarity.
We highlight that 98\% of the sentences in STS13 dataset consist of at most 48 tokens and particularly, the time difference is negligible in case of predicting the sentence pairs whose sentences have less than 48 tokens.
This result shows that significant increment of inference time does not occur within the range of the sequence length owing to parallel GPU computations, even though \proposeddist has the quadratic time complexity with respect to the sentence length.

\section{Conclusion}
\label{sec:conclusion}
In this work, we present \proposed, a learning framework for an interpretable sentence similarity model based on optimal transport.
First, we view each sentence similarity measure as a transportation problem, pointing out the unexpressiveness of the existing pooling-based similarity.
Integrating the concept of optimal transport into a pretrained language model, \proposed defines the distance measure by using the semantically-aligned token pairs between two sentences and furthermore, it finetunes a model with this distance based on contrastive learning for better interpretability.
We empirically show that \proposed accurately predicts sentence similarity while providing interpretable token pair contributions consistent with human judgements.
With the belief that the ability to interpret model behavior is critical for future AI models,
we focus on enhancing this virtue targeted on STS task throughout this research.

\section*{Acknowledgements}
This work was supported by Institute of Information \& communications Technology Planning \& Evaluation (IITP) grant funded by the Korea government (MSIT) (No.2018-0-00584, (SW starlab) Development of Decision Support System Software based on Next-Generation Machine Learning) and the NRF grant funded by the MSIT (South Korea, No.2020R1A2B5B03097210)
) and Institute of Information \& communications Technology Planning \& Evaluation (IITP) grant funded by the Korea government (MSIT) (No.2019-0-01906, Artificial Intelligence Graduate School Program(POSTECH)).

\bibliography{anthology,custom}

\begin{thebibliography}{39}
\expandafter\ifx\csname natexlab\endcsname\relax\def\natexlab#1{#1}\fi

\bibitem[{Agirre et~al.(2015)Agirre, Banea, Cardie, Cer, Diab, Gonzalez-Agirre,
  Guo, Lopez-Gazpio, Maritxalar, Mihalcea, Rigau, Uria, and
  Wiebe}]{agirre-etal-2015-semeval}
Eneko Agirre, Carmen Banea, Claire Cardie, Daniel Cer, Mona Diab, Aitor
  Gonzalez-Agirre, Weiwei Guo, I{\~n}igo Lopez-Gazpio, Montse Maritxalar, Rada
  Mihalcea, German Rigau, Larraitz Uria, and Janyce Wiebe. 2015.
\newblock \href {https://doi.org/10.18653/v1/S15-2045} {{S}em{E}val-2015 task
  2: Semantic textual similarity, {E}nglish, {S}panish and pilot on
  interpretability}.
\newblock In \emph{Proceedings of the 9th International Workshop on Semantic
  Evaluation ({S}em{E}val 2015)}, pages 252--263, Denver, Colorado. Association
  for Computational Linguistics.

\bibitem[{Agirre et~al.(2016)Agirre, Gonzalez-Agirre, Lopez-Gazpio, Maritxalar,
  Rigau, and Uria}]{agirre-etal-2016-semeval-2016}
Eneko Agirre, Aitor Gonzalez-Agirre, I{\~n}igo Lopez-Gazpio, Montse Maritxalar,
  German Rigau, and Larraitz Uria. 2016.
\newblock \href {https://doi.org/10.18653/v1/S16-1082} {{S}em{E}val-2016 task
  2: Interpretable semantic textual similarity}.
\newblock In \emph{Proceedings of the 10th International Workshop on Semantic
  Evaluation ({S}em{E}val-2016)}, pages 512--524, San Diego, California.
  Association for Computational Linguistics.

\bibitem[{Belinkov and Glass(2019)}]{belinkov-glass-2019-analysis}
Yonatan Belinkov and James Glass. 2019.
\newblock \href {https://doi.org/10.1162/tacl_a_00254} {Analysis methods in
  neural language processing: A survey}.
\newblock \emph{Transactions of the Association for Computational Linguistics},
  7:49--72.

\bibitem[{Bowman et~al.(2015)Bowman, Angeli, Potts, and
  Manning}]{bowman-etal-2015-large}
Samuel~R. Bowman, Gabor Angeli, Christopher Potts, and Christopher~D. Manning.
  2015.
\newblock \href {https://doi.org/10.18653/v1/D15-1075} {A large annotated
  corpus for learning natural language inference}.
\newblock In \emph{Proceedings of the 2015 Conference on Empirical Methods in
  Natural Language Processing}, pages 632--642, Lisbon, Portugal. Association
  for Computational Linguistics.

\bibitem[{Cer et~al.(2017)Cer, Diab, Agirre, Lopez-Gazpio, and
  Specia}]{cer-etal-2017-semeval}
Daniel Cer, Mona Diab, Eneko Agirre, I{\~n}igo Lopez-Gazpio, and Lucia Specia.
  2017.
\newblock \href {https://doi.org/10.18653/v1/S17-2001} {{S}em{E}val-2017 task
  1: Semantic textual similarity multilingual and crosslingual focused
  evaluation}.
\newblock In \emph{Proceedings of the 11th International Workshop on Semantic
  Evaluation ({S}em{E}val-2017)}, pages 1--14, Vancouver, Canada. Association
  for Computational Linguistics.

\bibitem[{Chen et~al.(2020)Chen, Kornblith, Norouzi, and
  Hinton}]{pmlr-v119-chen20j}
Ting Chen, Simon Kornblith, Mohammad Norouzi, and Geoffrey Hinton. 2020.
\newblock \href {https://proceedings.mlr.press/v119/chen20j.html} {A simple
  framework for contrastive learning of visual representations}.
\newblock In \emph{Proceedings of the 37th International Conference on Machine
  Learning}, volume 119 of \emph{Proceedings of Machine Learning Research},
  pages 1597--1607. PMLR.

\bibitem[{Conneau and Kiela(2018)}]{conneau-kiela-2018-senteval}
Alexis Conneau and Douwe Kiela. 2018.
\newblock \href {https://aclanthology.org/L18-1269} {{S}ent{E}val: An
  evaluation toolkit for universal sentence representations}.
\newblock In \emph{Proceedings of the Eleventh International Conference on
  Language Resources and Evaluation ({LREC} 2018)}, Miyazaki, Japan. European
  Language Resources Association (ELRA).

\bibitem[{Cuturi(2013)}]{NIPS2013_af21d0c9}
Marco Cuturi. 2013.
\newblock \href
  {https://proceedings.neurips.cc/paper/2013/file/af21d0c97db2e27e13572cbf59eb343d-Paper.pdf}
  {Sinkhorn distances: Lightspeed computation of optimal transport}.
\newblock In \emph{Advances in Neural Information Processing Systems},
  volume~26. Curran Associates, Inc.

\bibitem[{Devlin et~al.(2019)Devlin, Chang, Lee, and
  Toutanova}]{devlin-etal-2019-bert}
Jacob Devlin, Ming-Wei Chang, Kenton Lee, and Kristina Toutanova. 2019.
\newblock \href {https://doi.org/10.18653/v1/N19-1423} {{BERT}: Pre-training of
  deep bidirectional transformers for language understanding}.
\newblock In \emph{Proceedings of the 2019 Conference of the North {A}merican
  Chapter of the Association for Computational Linguistics: Human Language
  Technologies, Volume 1 (Long and Short Papers)}, pages 4171--4186,
  Minneapolis, Minnesota. Association for Computational Linguistics.

\bibitem[{Gao et~al.(2021)Gao, Yao, and Chen}]{gao2021simcse}
Tianyu Gao, Xingcheng Yao, and Danqi Chen. 2021.
\newblock Simcse: Simple contrastive learning of sentence embeddings.
\newblock \emph{arXiv preprint arXiv:2104.08821}.

\bibitem[{Gilpin et~al.(2018)Gilpin, Bau, Yuan, Bajwa, Specter, and
  Kagal}]{gilpin2018explaining}
Leilani~H Gilpin, David Bau, Ben~Z Yuan, Ayesha Bajwa, Michael Specter, and
  Lalana Kagal. 2018.
\newblock Explaining explanations: An overview of interpretability of machine
  learning.
\newblock In \emph{2018 IEEE 5th International Conference on data science and
  advanced analytics (DSAA)}, pages 80--89. IEEE.

\bibitem[{Gomaa et~al.(2013)Gomaa, Fahmy et~al.}]{gomaa2013survey}
Wael~H Gomaa, Aly~A Fahmy, et~al. 2013.
\newblock A survey of text similarity approaches.
\newblock \emph{international journal of Computer Applications}, 68(13):13--18.

\bibitem[{Grauman and Darrell(2004)}]{grauman2004fast}
Kristen Grauman and Trevor Darrell. 2004.
\newblock Fast contour matching using approximate earth mover's distance.
\newblock In \emph{Proceedings of the 2004 IEEE Computer Society Conference on
  Computer Vision and Pattern Recognition, 2004. CVPR 2004.}, volume~1, pages
  I--I. IEEE.

\bibitem[{Humeau et~al.(2020)Humeau, Shuster, Lachaux, and
  Weston}]{Humeau2020Poly-encoders:}
Samuel Humeau, Kurt Shuster, Marie-Anne Lachaux, and Jason Weston. 2020.
\newblock \href {https://openreview.net/forum?id=SkxgnnNFvH} {Poly-encoders:
  Architectures and pre-training strategies for fast and accurate
  multi-sentence scoring}.
\newblock In \emph{International Conference on Learning Representations}.

\bibitem[{Konop{\'\i}k et~al.(2016)Konop{\'\i}k, Pra{\v{z}}{\'a}k, Steinberger,
  and Brychc{\'\i}n}]{konopik-etal-2016-uwb}
Miloslav Konop{\'\i}k, Ond{\v{r}}ej Pra{\v{z}}{\'a}k, David Steinberger, and
  Tom{\'a}{\v{s}} Brychc{\'\i}n. 2016.
\newblock \href {https://doi.org/10.18653/v1/S16-1124} {{UWB} at
  {S}em{E}val-2016 task 2: Interpretable semantic textual similarity with
  distributional semantics for chunks}.
\newblock In \emph{Proceedings of the 10th International Workshop on Semantic
  Evaluation ({S}em{E}val-2016)}, pages 803--808, San Diego, California.
  Association for Computational Linguistics.

\bibitem[{Kusner et~al.(2015)Kusner, Sun, Kolkin, and
  Weinberger}]{pmlr-v37-kusnerb15}
Matt Kusner, Yu~Sun, Nicholas Kolkin, and Kilian Weinberger. 2015.
\newblock \href {https://proceedings.mlr.press/v37/kusnerb15.html} {From word
  embeddings to document distances}.
\newblock In \emph{Proceedings of the 32nd International Conference on Machine
  Learning}, volume~37 of \emph{Proceedings of Machine Learning Research},
  pages 957--966, Lille, France. PMLR.

\bibitem[{Lee et~al.(2021{\natexlab{a}})Lee, Kim, Lee, Park, and Yu}]{9679111}
Dongha Lee, Su~Kim, Seonghyeon Lee, Chanyoung Park, and Hwanjo Yu.
  2021{\natexlab{a}}.
\newblock \href {https://doi.org/10.1109/ICDM51629.2021.00142} {Learnable
  structural semantic readout for graph classification}.
\newblock In \emph{2021 IEEE International Conference on Data Mining (ICDM)},
  pages 1180--1185.

\bibitem[{Lee et~al.(2021{\natexlab{b}})Lee, Lee, and Yu}]{LeeLY21}
Dongha Lee, Seonghyeon Lee, and Hwanjo Yu. 2021{\natexlab{b}}.
\newblock \href {https://ojs.aaai.org/index.php/AAAI/article/view/17008}
  {Learnable dynamic temporal pooling for time series classification}.
\newblock In \emph{AAAI}, pages 8288--8296.

\bibitem[{Li et~al.(2020)Li, Li, Wang, Fu, Lin, Chen, Zhang, Tao, Zhang, Wang,
  Shen, Yang, and Carin}]{li-etal-2020-improving-text}
Jianqiao Li, Chunyuan Li, Guoyin Wang, Hao Fu, Yuhchen Lin, Liqun Chen, Yizhe
  Zhang, Chenyang Tao, Ruiyi Zhang, Wenlin Wang, Dinghan Shen, Qian Yang, and
  Lawrence Carin. 2020.
\newblock \href {https://doi.org/10.18653/v1/2020.emnlp-main.735} {Improving
  text generation with student-forcing optimal transport}.
\newblock In \emph{Proceedings of the 2020 Conference on Empirical Methods in
  Natural Language Processing (EMNLP)}, pages 9144--9156, Online. Association
  for Computational Linguistics.

\bibitem[{Liu et~al.(2019)Liu, Ott, Goyal, Du, Joshi, Chen, Levy, Lewis,
  Zettlemoyer, and Stoyanov}]{liu2019roberta}
Yinhan Liu, Myle Ott, Naman Goyal, Jingfei Du, Mandar Joshi, Danqi Chen, Omer
  Levy, Mike Lewis, Luke Zettlemoyer, and Veselin Stoyanov. 2019.
\newblock Roberta: A robustly optimized bert pretraining approach.
\newblock \emph{arXiv preprint arXiv:1907.11692}.

\bibitem[{Lopez-Gazpio et~al.(2016)Lopez-Gazpio, Agirre, and
  Maritxalar}]{lopez-gazpio-etal-2016-iubc}
I{\~n}igo Lopez-Gazpio, Eneko Agirre, and Montse Maritxalar. 2016.
\newblock \href {https://doi.org/10.18653/v1/S16-1119} {i{UBC} at
  {S}em{E}val-2016 task 2: {RNN}s and {LSTM}s for interpretable {STS}}.
\newblock In \emph{Proceedings of the 10th International Workshop on Semantic
  Evaluation ({S}em{E}val-2016)}, pages 771--776, San Diego, California.
  Association for Computational Linguistics.

\bibitem[{Maji et~al.(2020)Maji, Kumar, Bansal, Roy, and
  Goyal}]{ijcai2020-0333}
Subhadeep Maji, Rohan Kumar, Manish Bansal, Kalyani Roy, and Pawan Goyal. 2020.
\newblock \href {https://doi.org/10.24963/ijcai.2020/333} {Logic constrained
  pointer networks for interpretable textual similarity}.
\newblock In \emph{Proceedings of the Twenty-Ninth International Joint
  Conference on Artificial Intelligence, {IJCAI-20}}, pages 2405--2411.
  International Joint Conferences on Artificial Intelligence Organization.
\newblock Main track.

\bibitem[{Majumder et~al.(2016)Majumder, Pakray, Gelbukh, and
  Pinto}]{majumder2016semantic}
Goutam Majumder, Partha Pakray, Alexander Gelbukh, and David Pinto. 2016.
\newblock Semantic textual similarity methods, tools, and applications: A
  survey.
\newblock \emph{Computaci{\'o}n y Sistemas}, 20(4):647--665.

\bibitem[{Monge(1781)}]{monge1781memoire}
Gaspard Monge. 1781.
\newblock M{\'e}moire sur la th{\'e}orie des d{\'e}blais et des remblais.
\newblock \emph{Histoire de l'Acad{\'e}mie Royale des Sciences de Paris}.

\bibitem[{Reimers and Gurevych(2019)}]{reimers-gurevych-2019-sentence}
Nils Reimers and Iryna Gurevych. 2019.
\newblock \href {https://doi.org/10.18653/v1/D19-1410} {Sentence-{BERT}:
  Sentence embeddings using {S}iamese {BERT}-networks}.
\newblock In \emph{Proceedings of the 2019 Conference on Empirical Methods in
  Natural Language Processing and the 9th International Joint Conference on
  Natural Language Processing (EMNLP-IJCNLP)}, pages 3982--3992, Hong Kong,
  China. Association for Computational Linguistics.

\bibitem[{Rogers et~al.(2020)Rogers, Kovaleva, and
  Rumshisky}]{rogers-etal-2020-primer}
Anna Rogers, Olga Kovaleva, and Anna Rumshisky. 2020.
\newblock \href {https://doi.org/10.1162/tacl_a_00349} {A primer in
  {BERT}ology: What we know about how {BERT} works}.
\newblock \emph{Transactions of the Association for Computational Linguistics},
  8:842--866.

\bibitem[{Salton and Buckley(1988)}]{salton1988term}
Gerard Salton and Christopher Buckley. 1988.
\newblock Term-weighting approaches in automatic text retrieval.
\newblock \emph{Information processing \& management}, 24(5):513--523.

\bibitem[{Shirdhonkar and Jacobs(2008)}]{shirdhonkar2008approximate}
Sameer Shirdhonkar and David~W Jacobs. 2008.
\newblock Approximate earth mover’s distance in linear time.
\newblock In \emph{2008 IEEE Conference on Computer Vision and Pattern
  Recognition}, pages 1--8. IEEE.

\bibitem[{Sultan et~al.(2015)Sultan, Bethard, and
  Sumner}]{sultan-etal-2015-dls}
Md~Arafat Sultan, Steven Bethard, and Tamara Sumner. 2015.
\newblock \href {https://doi.org/10.18653/v1/S15-2027} {{DLS}@{CU}: Sentence
  similarity from word alignment and semantic vector composition}.
\newblock In \emph{Proceedings of the 9th International Workshop on Semantic
  Evaluation ({S}em{E}val 2015)}, pages 148--153, Denver, Colorado. Association
  for Computational Linguistics.

\bibitem[{Swanson et~al.(2020)Swanson, Yu, and
  Lei}]{swanson-etal-2020-rationalizing}
Kyle Swanson, Lili Yu, and Tao Lei. 2020.
\newblock \href {https://doi.org/10.18653/v1/2020.acl-main.496} {Rationalizing
  text matching: {L}earning sparse alignments via optimal transport}.
\newblock In \emph{Proceedings of the 58th Annual Meeting of the Association
  for Computational Linguistics}, pages 5609--5626, Online. Association for
  Computational Linguistics.

\bibitem[{Tekumalla and Jat(2016)}]{tekumalla-jat-2016-iiscnlp}
Lavanya Tekumalla and Sharmistha Jat. 2016.
\newblock \href {https://doi.org/10.18653/v1/S16-1122} {{IISCNLP} at
  {S}em{E}val-2016 task 2: Interpretable {STS} with {ILP} based multiple chunk
  aligner}.
\newblock In \emph{Proceedings of the 10th International Workshop on Semantic
  Evaluation ({S}em{E}val-2016)}, pages 790--795, San Diego, California.
  Association for Computational Linguistics.

\bibitem[{Villani(2008)}]{villani2008optimal}
C{\'e}dric Villani. 2008.
\newblock \emph{Optimal Transport: Old and New}, volume 338.
\newblock Springer Science \& Business Media.

\bibitem[{Williams et~al.(2018)Williams, Nangia, and
  Bowman}]{williams-etal-2018-broad}
Adina Williams, Nikita Nangia, and Samuel Bowman. 2018.
\newblock \href {https://doi.org/10.18653/v1/N18-1101} {A broad-coverage
  challenge corpus for sentence understanding through inference}.
\newblock In \emph{Proceedings of the 2018 Conference of the North {A}merican
  Chapter of the Association for Computational Linguistics: Human Language
  Technologies, Volume 1 (Long Papers)}, pages 1112--1122, New Orleans,
  Louisiana. Association for Computational Linguistics.

\bibitem[{Wu et~al.(2018)Wu, Yen, Xu, Xu, Balakrishnan, Chen, Ravikumar, and
  Witbrock}]{wu-etal-2018-word}
Lingfei Wu, Ian En-Hsu Yen, Kun Xu, Fangli Xu, Avinash Balakrishnan, Pin-Yu
  Chen, Pradeep Ravikumar, and Michael~J. Witbrock. 2018.
\newblock \href {https://doi.org/10.18653/v1/D18-1482} {Word mover{'}s
  embedding: From {W}ord2{V}ec to document embedding}.
\newblock In \emph{Proceedings of the 2018 Conference on Empirical Methods in
  Natural Language Processing}, pages 4524--4534, Brussels, Belgium.
  Association for Computational Linguistics.

\bibitem[{Wu et~al.(2020)Wu, Wang, Gu, Khabsa, Sun, and Ma}]{wu2020clear}
Zhuofeng Wu, Sinong Wang, Jiatao Gu, Madian Khabsa, Fei Sun, and Hao Ma. 2020.
\newblock Clear: Contrastive learning for sentence representation.
\newblock \emph{arXiv preprint arXiv:2012.15466}.

\bibitem[{Xu et~al.(2021)Xu, Zhou, Gan, Zheng, and
  Li}]{xu-etal-2021-vocabulary}
Jingjing Xu, Hao Zhou, Chun Gan, Zaixiang Zheng, and Lei Li. 2021.
\newblock \href {https://doi.org/10.18653/v1/2021.acl-long.571} {Vocabulary
  learning via optimal transport for neural machine translation}.
\newblock In \emph{Proceedings of the 59th Annual Meeting of the Association
  for Computational Linguistics and the 11th International Joint Conference on
  Natural Language Processing (Volume 1: Long Papers)}, pages 7361--7373,
  Online. Association for Computational Linguistics.

\bibitem[{Yan et~al.(2021)Yan, Li, Wang, Zhang, Wu, and
  Xu}]{yan-etal-2021-consert}
Yuanmeng Yan, Rumei Li, Sirui Wang, Fuzheng Zhang, Wei Wu, and Weiran Xu. 2021.
\newblock \href {https://doi.org/10.18653/v1/2021.acl-long.393} {{C}on{SERT}: A
  contrastive framework for self-supervised sentence representation transfer}.
\newblock In \emph{Proceedings of the 59th Annual Meeting of the Association
  for Computational Linguistics and the 11th International Joint Conference on
  Natural Language Processing (Volume 1: Long Papers)}, pages 5065--5075,
  Online. Association for Computational Linguistics.

\bibitem[{Zhang et~al.(2020)Zhang, Kishore, Wu, Weinberger, and
  Artzi}]{Zhang*2020BERTScore:}
Tianyi Zhang, Varsha Kishore, Felix Wu, Kilian~Q. Weinberger, and Yoav Artzi.
  2020.
\newblock \href {https://openreview.net/forum?id=SkeHuCVFDr} {Bertscore:
  Evaluating text generation with bert}.
\newblock In \emph{International Conference on Learning Representations}.

\bibitem[{Zhao et~al.(2019)Zhao, Peyrard, Liu, Gao, Meyer, and
  Eger}]{zhao-etal-2019-moverscore}
Wei Zhao, Maxime Peyrard, Fei Liu, Yang Gao, Christian~M. Meyer, and Steffen
  Eger. 2019.
\newblock \href {https://doi.org/10.18653/v1/D19-1053} {{M}over{S}core: Text
  generation evaluating with contextualized embeddings and earth mover
  distance}.
\newblock In \emph{Proceedings of the 2019 Conference on Empirical Methods in
  Natural Language Processing and the 9th International Joint Conference on
  Natural Language Processing (EMNLP-IJCNLP)}, pages 563--578, Hong Kong,
  China. Association for Computational Linguistics.

\end{thebibliography}
\bibliographystyle{acl_natbib}




\end{document}